\documentclass[letterpaper]{article} % DO NOT CHANGE THIS
\usepackage[preprint]{aaai2027}  % DO NOT CHANGE THIS
\usepackage[hyphens]{url}  % DO NOT CHANGE THIS
\usepackage{graphicx} % DO NOT CHANGE THIS
\usepackage{natbib}  % DO NOT CHANGE THIS AND DO NOT ADD ANY OPTIONS TO IT
\usepackage{caption} % DO NOT CHANGE THIS AND DO NOT ADD ANY OPTIONS TO IT
\usepackage{algorithm}
\usepackage{algorithmic}

\usepackage{newfloat}
\usepackage{listings}
\DeclareCaptionStyle{ruled}{labelfont=normalfont,labelsep=colon,strut=off} % DO NOT CHANGE THIS
\floatstyle{ruled}
\newfloat{listing}{tb}{lst}{}
\floatname{listing}{Listing}

\usepackage{booktabs}
 \usepackage{colortbl}
\usepackage{amsmath}
\usepackage{amssymb}

\definecolor{ourrow}{HTML}{F4F4F4}
\definecolor{gainclr}{HTML}{1F8A4C}
\newcommand{\ours}{\textsc{Ours}}
\newcommand{\best}[1]{\textbf{#1}}
\newcommand{\snd}[1]{\underline{#1}}
\newcommand{\gain}[1]{{\scriptsize$\uparrow$#1}}
\newcommand{\na}{$-$}

\title{Hidden in Plain Sight: Diffusion-Based Unrestricted Robotic Attacks on Vision-Language-Action Models}
\author{
    Jiahui Han\textsuperscript{\rm 1,2,*},
    Yuhui Yao\textsuperscript{\rm 2,3,*},
    Xin Wang\textsuperscript{\rm 2,*,\textdagger},
    Jiafei Cao\textsuperscript{\rm 2},
    Mingxuan Zhang\textsuperscript{\rm 2},
    Danfeng Shan\textsuperscript{\rm 1},\\
    Huiqi Deng\textsuperscript{\rm 1,2,\textdagger},
    Guanchu Wang\textsuperscript{\rm 2,\textdagger},
    Xia Hu\textsuperscript{\rm 2,\textdagger}
}

\affiliations{
    \textsuperscript{\rm 1}Xi'an Jiaotong University\\
    \textsuperscript{\rm 2}Shanghai AI Laboratory\\
    \textsuperscript{\rm 3}University of Science and Technology of China
}

\begin{document}

\maketitle

\begingroup
\makeatletter
\renewcommand{\thefootnote}{}
\renewcommand{\@makefntext}[1]{\noindent #1}
\footnotetext{
\textsuperscript{*} Equal contribution.\\
\textsuperscript{\textdagger} Corresponding authors: Huiqi Deng, Guanchu Wang, Xin Wang, and Xia Hu.
}
\makeatother
\endgroup

\maketitle

\begin{abstract}
% AAAI creates proceedings, working notes, and technical reports directly from electronic source furnished by the authors. To ensure that all papers in the publication have a uniform appearance, authors must adhere to the following instructions.
Vision-Language-Action (VLA) models have shown strong capabilities in controlling robots across diverse manipulation tasks. However, their adversarial robustness remains largely underexplored, and exploiting this weakness can lead to physical-world harm.
% Existing attacks mainly perturb pixels and require white-box access to the victim model. The resulting adversarial inputs are unnatural and easy to detect, which makes them impractical to deploy in real physical environments. 
Existing attacks on VLA models often rely on pixel-space perturbations or white-box access, resulting in noticeable artifacts and limited deployability in real-world robotic systems.
% In this work, we propose \textbf{DURA}, a diffusion-based unrestricted robotic attack that generates visually natural adversarial patches for VLA models. DURA searches along the latent trajectory of a pretrained diffusion model and steers the robot toward an attacker-specified target action, while keeping the patch consistent with natural scene content. DURA supports both white-box and black-box settings, where the black-box setting requires only the predicted actions of the victim model. 
In this work, we propose DURA, a diffusion-based unrestricted robotic attack that generates visually natural adversarial patches for VLA models. DURA supports both white-box and black-box attack settings, where the black-box setting requires only the predicted actions of the victim model. By optimizing along the latent trajectory of a pretrained diffusion model, DURA generates visually natural patches while steering the robot toward attacker-specified target actions.
Extensive experiments in both simulation and the real physical world show that DURA consistently outperforms existing methods. Our findings expose a safety risk for physically deployed VLA models and call for stronger defenses.
\end{abstract}


\section{Introduction}
% Vision-Language-Action (VLA) models map visual observations, proprioceptive state, and language instructions directly to robot actions through a single end-to-end policy~\citep{kim2024openvla,qu2025spatialvla,black2024pi_0,team2024octo}. They originate from the integration of Large Language Models and Vision-Language Models with imitation-learning robot control; early systems such as RT-1~\citep{brohan2022rt} and RT-2~\citep{zitkovich2023rt} first demonstrated the feasibility of foundation-model-driven robot policies. Subsequent open-source releases such as OpenVLA~\citep{kim2024openvla} and Octo~\citep{team2024octo} scaled the paradigm to cross-embodiment generalist policies trained on diverse manipulation data. More recent systems such as $\pi_0$~\citep{black2024pi_0} and SpatialVLA~\citep{qu2025spatialvla} have advanced flow-matching action experts and 3D-aware spatial reasoning. VLA models now power a wide range of real-world robotic systems, and industry is being explored for commercial deployment.
Vision-Language-Action (VLA) models have become a powerful paradigm for general-purpose robotic control~\citep{kim2024openvla}, unifying visual perception, language understanding, and action prediction within a single policy. 
Given an image observation, robot state, and language instruction, a VLA model can directly generate actions for open vocabulary manipulation tasks.
Representative systems, from RT-1~\citep{brohan2022rt} and RT-2~\citep{zitkovich2023rt} to recent OpenVLA~\citep{kim2024openvla}, Octo~\citep{team2024octo}, $\pi_0$~\citep{black2024pi_0}, and SpatialVLA~\citep{qu2025spatialvla} have demonstrated increasing generalization across tasks, embodiments, and action representations. 
These advances point toward generalist robotic agents that can follow diverse instructions across complex environments, bringing VLA models increasingly closer to deployment in open-world robotic systems.\looseness=-1
% As these VLA models gain greater autonomy in interpreting instructions and executing actions, ensuring that they behave reliably under unexpected or adversarial inputsbecomes essential for safe deployment.~\citep{wang2024advqdet,wang2025tapt,zhang2026safevla,wang2026tame} 

\begin{figure}[t]
  \centering
  \makebox[\linewidth][c]{%
    \includegraphics[width=\linewidth]{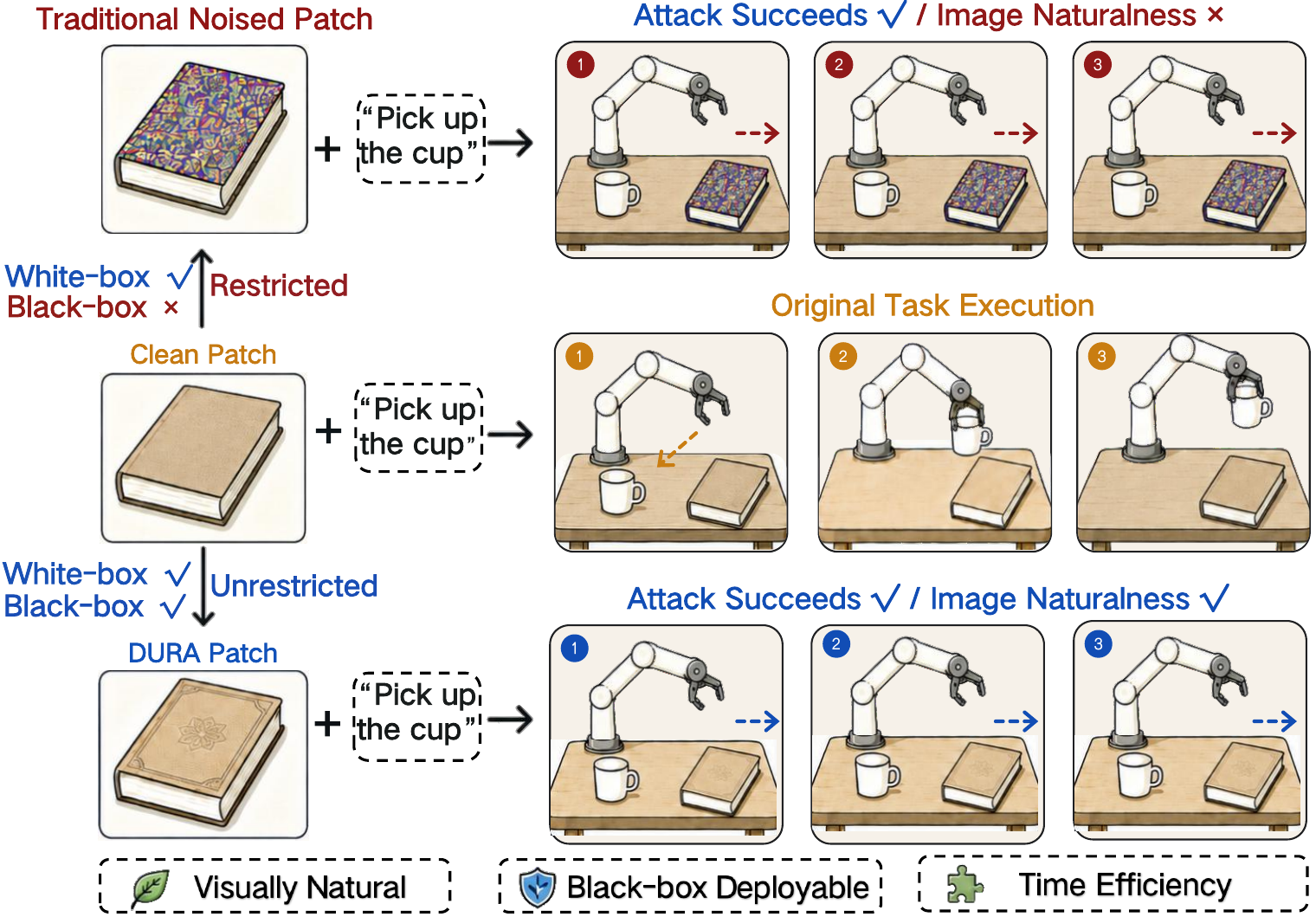}%
  }
  \caption{An illustration of unrestricted patch attacks on a VLA robot (``pick up the
  cup''). \textbf{Top:} a traditional noised patch stalls the arm but is conspicuous and
  white-box only. \textbf{Bottom:} our DURA patch stalls the arm while being
  \emph{visually natural}, \emph{black-box deployable}, and \emph{semantically
  consistent}.}
  \label{fig:fig_1}
\end{figure}

% Recent studies show that VLA models are vulnerable to adversarial attacks that target both image and text input \citep{kim2024openvla,zhang2025safevla,wang2025exploring,jones2025adversarial}. Wang et al.~\citep{wang2025exploring} design motion-aware patch objectives, while FreezeVLA~\citep{wang2025freezevla} instead targets inaction itself, using a max-min bi-level objective whose images freeze robots across diverse prompts. However, these attacks typically optimize their patches directly at the pixel level under tight perturbation budgets. 
% The resulting patterns either appear as obvious noise that no human supervisor would overlook, or remain confined to a narrow constrained space that cannot accommodate richer, scene-consistent textures. On the threat-model side, existing attack methods assume full white-box access to the target VLA model, allowing direct readout of model parameters and gradient backpropagation. This assumption rarely holds in real deployment, where commercial robotic systems expose only an inference interface and treat their VLA backbone and training data as trade secrets. 
% Despite their strong capabilities, VLA models remain vulnerable to adversarial inputs
However, the growing autonomy of VLA models also introduces substantial safety risks~\citep{wang2024advqdet,wang2025tapt,wang2026openrt,ma2026safety}. Recent studies show that carefully crafted visual patches or malicious language instructions can disrupt robot execution, causing the policy to freeze, deviate from the intended task, or follow target behaviors~\citep{zhang2026safevla,wang2025exploring,jones2025adversarial,wang2025freezevla}. While these attacks provide valuable evidence of VLA safety risks, they still fall short of realistic deployment conditions. Specifically, (1) \textbf{Visual naturalness:} existing visual patches often appear as meaningless or conspicuous artifacts rather than natural scene content, making them easy to detect in real workspaces; (2) \textbf{Black-box deployability:} existing VLA attacks commonly rely on privileged internal access, such as model parameters, gradients, or logits, which is rarely available for closed-source robotic systems; (3) \textbf{Time efficiency:} they often require lengthy per-instance optimization to craft a single patch, which incurs a high query or compute cost and limits their use in practice. These limitations shift the central question from demonstrating white-box vulnerabilities to constructing visually natural, black-box adversarial perturbations with real-world effectiveness.

In this work, we propose \textbf{DURA}, a diffusion-based unrestricted robotic attack for VLA models, as illustrated in Figure~\ref{fig:fig_1}. Rather than crafting small norm-bounded perturbations or high-frequency additive noise on a fixed patch, DURA treats the localized patch content as unrestricted and searches along the latent trajectory of a pretrained diffusion model, using the diffusion prior to preserve visual naturalness while the VLA attack objective steers the robot toward an attacker-specified target action. 
% Specifically, DURA injects the VLA attack update into the DDIM reverse process and anchors the adversarial trajectory to a clean diffusion path, stabilizing the latent optimization and keeping the generated patch close to the natural-image manifold.
% The same trajectory supports both white-box and black-box threat models. Under white-box access, DURA backpropagates the attack loss through the VLA policy and the VAE decoder\citep{kingma2013auto}. Under black-box access, DURA estimates latent updates from action-output queries without accessing model parameters or gradients. By optimizing a single patch over diverse frames and instructions, DURA further improves transferability across robotic tasks and deployment conditions.

The resulting latent-trajectory framework naturally accommodates both white-box and black-box threat models, differing only in how the per-step attack direction is estimated. Under white-box access, DURA computes the latent update by backpropagating the target-action loss through the VLA policy and the VAE decoder~\citep{kingma2013auto}. Under black-box access, it estimates the update from action-output queries alone, requiring no access to model parameters or gradients. By optimizing a single patch over diverse frames and instructions, DURA further improves transferability across robotic tasks and deployment conditions.

We evaluate DURA on two SOTA open-source VLA models, OpenVLA~\citep{kim2024openvla} and $\pi_{0}$-FAST~\citep{karl2025fast}, across both the LIBERO simulation benchmark~\citep{liu2023libero} and a real Franka robot arm. Our results show that DURA achieves substantially higher attack success rates (ASR) than existing baselines under black-box access, while remaining effective under common input-transformation defenses and transferring to the real Franka arm as a controllable, on-demand switch over robot behavior. These results highlight the urgent need to assess and mitigate deployability-level vulnerabilities of current VLA models.

% In summary, our main contributions are:
% \begin{itemize}
%     \item We propose \textbf{DURA}, a diffusion-based unrestricted adversarial patch attack for VLA models. Unlike meaningless perturbations, DURA generates semantically meaningful and visually natural patches that steer robot policies toward attacker-specified actions.

%     \item We formulate DURA as a unified latent-space optimization framework along the DDIM reverse trajectory of a pretrained diffusion model. By anchoring the adversarial trajectory to a clean denoising path, DURA preserves visual naturalness while supporting both white-box updates through local gradient backpropagation and black-box updates through Tweedie-based score-function estimation.In addition, DURA achieves higher optimization efficiency than pixel-space baselines in both white-box and black-box settings.
%     % In addition, DURA achieves both the white-box and black-box settings with higher time efficiency.

%     \item We conduct extensive evaluations on OpenVLA and $\pi_{0}$-FAST across the LIBERO benchmark and a real Franka robot arm. Under black-box access, DURA achieves average ASRs of 86.0\% in simulation and 79.3\% in the physical setting, outperforming the baseline by 43.8 and 40.1 percentage points, respectively.
% \end{itemize}

In summary, our main contributions are:

\begin{itemize}

\item We propose \textbf{DURA}, a diffusion-based unrestricted robotic attack for Vision-Language-Action models. DURA generates localized, visually natural, and semantically plausible patches that steer robot policies toward attacker-specified actions.

\item We establish DURA as a practical attack framework for both white-box and action-output black-box settings. By searching in a diffusion-guided patch space, DURA achieves a favorable trade-off among attack effectiveness, visual naturalness, and optimization efficiency.

% \item We conduct comprehensive evaluations on OpenVLA and $\pi_{0}$-FAST across LIBERO and real-robot Franka experiments. DURA consistently outperforms strong pixel-space and query-based baselines in attack success, target-action control, visual naturalness, and time efficiency. On OpenVLA, DURA reaches 100\% ASR under white-box access in both the Simulated and Physical settings; under black-box access, it achieves average ASRs of 86.0\% and 79.3\% in the two settings, improving over the strongest baseline by 43.8 and 40.1 percentage points, respectively.
\item We conduct comprehensive evaluations on OpenVLA and $\pi_{0}$-FAST across LIBERO and a real-world Franka robot. DURA achieves \textbf{79.3–100\%} ASR across the evaluated white-box and black-box settings, outperforming the strongest target-only baselines. Real-robot experiments further demonstrate controllable and repeatable targeted behaviors on physical hardware.

\end{itemize}

\section{Related Work}
\label{sec:Related Work}

\noindent\textbf{Adversarial Attacks on Vision-Language-Action Models.}
Existing attacks on VLA models target both visual and language inputs.
On the visual channel, Wang et al.~\citep{wang2025exploring} study
patch-based objectives for disrupting or steering robot actions, while
FreezeVLA~\citep{wang2025freezevla} uses bi-level optimization to
induce action freezing. These methods achieve strong attack performance
but require white-box access to the victim model.

Transferable attacks reduce this requirement by optimizing patches on
accessible models or shared representations. Representative methods
include UPA-RFAS~\citep{lu2026robots}, EDPA~\citep{xu2025model},
VLA-Hijack~\citep{fu2026vla}, and TRAP~\citep{huang2026trap}.
Although they avoid victim-model gradients at deployment, their
effectiveness relies on surrogate-to-victim transfer and may degrade
across different architectures and action representations. Across these
visual attack paradigms, patch naturalness has received limited
attention compared with attack success and transferability.

On the language channel, adversarial instructions can derail robot
execution~\citep{jones2025adversarial}, while
SABER~\citep{wu2026saber} performs agentic black-box attacks through
bounded instruction edits.

DURA provides a unified framework for both white-box and
black-box visual attacks. DURA uses gradients when available,
while estimating the attack direction directly from predicted actions
in the black-box setting. In both settings, a diffusion prior guides the optimization toward
natural-looking patches while inducing attacker-specified actions.

% \noindent\textbf{Diffusion-Based Unrestricted Adversarial Generation.???}
% Diffusion models generate images by progressively denoising a noised sample. Deterministic samplers such as DDIM~\citep{song2021denoising}, latent-space formulations~\citep{rombach2022high}, and score-based stochastic differential equations~\citep{song2020score} make this process efficient and controllable. Beyond image generation, a recent line of work uses pretrained diffusion models as priors for adversarial generation. Several recent works inject adversarial guidance into the reverse diffusion process to create natural and unrestricted adversarial examples, rather than adding bounded pixel noise~\citep{chen2023advdiffuser,gao2024retome,lin2025diffusion,chen2023natural}. AdvDiffVLM~\citep{guo2024efficient} brings this idea to vision-language models, embedding targeted adversarial semantics into the sampling process to produce transferable image-to-text attacks. Other works refine how the diffusion prior is exploited for adversarial generation\citep{chen2023content,dai2025semdiff,xue2023diffusion}. Diffusion priors have therefore become an effective tool for producing unrestricted adversarial content that remains visually natural. DURA extends this diffusion-guided unrestricted generation paradigm to VLA policies, where adversarial content must be optimized as a localized physical patch under action-level robotic feedback.
\noindent\textbf{Diffusion Priors for Natural Adversarial Examples.}
Diffusion models provide strong image priors for generating natural
adversarial examples beyond norm-bounded perturbations
~\citep{song2021denoising,rombach2022high,song2020score}.
Prior work incorporates adversarial guidance into reverse diffusion
~\citep{chen2023advdiffuser,gao2024retome,lin2025diffusion,
chen2023natural,chen2023content,dai2025semdiff,xue2023diffusion}.
AdvDiffVLM~\citep{guo2024efficient}, for example, embeds target
semantics into generated images to attack vision-language models.

Such targets can be expressed directly through image content: a class
or textual concept has a corresponding visual semantics. A target robot
action has no analogous visual form, as its relation to the image
depends on the instruction, scene, and robot state. Success on a single
frame is also insufficient because these factors evolve throughout
execution. DURA therefore optimizes one diffusion-constrained patch
jointly over diverse frames, coupling it to a common action objective
to induce consistent target behavior while retaining a natural
appearance.                                                                               
%===============================================================================

\section{Method}
  \begin{figure*}[t] 
  \centering 
  \includegraphics[width=\textwidth]{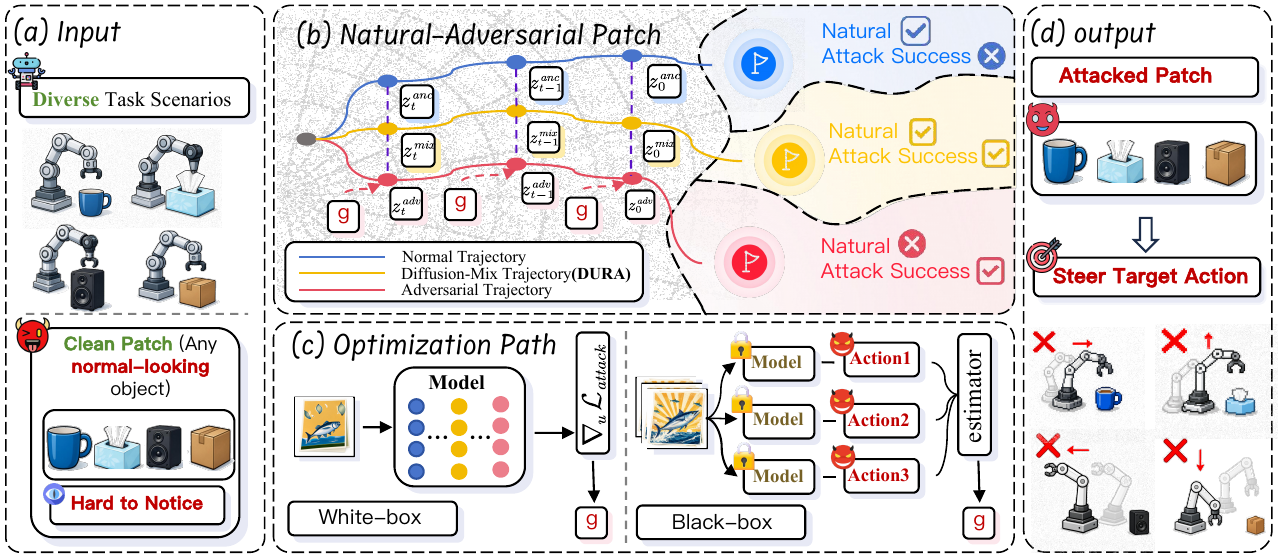}

\caption{The overall framework of DURA, which consists of (a) constructing diverse task scenarios with normal-looking clean patches, (b) generating natural-looking adversarial patches through diffusion-mixed trajectories, (c) optimizing the patches in either white-box or black-box settings, and (d) using the resulting adversarial patches to steer the robot toward attacker-specified target actions.}

  \label{fig:fig_main}
  \end{figure*}

\subsection{Threat Model}
We study targeted physical adversarial patch attacks against vision-language-action (VLA) models. The attacker can place a single localized patch in the robot's field of view, but cannot modify the model parameters, training data, robot state, or language instruction at test time. The patch is physically realizable and restricted to a limited image region, while its content is otherwise unrestricted. The attack goal is to drive the victim policy toward an attacker-specified target action whenever the patch is visible. We consider both white-box access, where gradients of the victim policy are available, and black-box action-output access, where the attacker can submit observations and language instructions to the victim policy and observe only the predicted action sequence. In both cases, we optimize a single patch over diverse observations to improve transfer across scenes and tasks.

\subsection{Problem Formulation} Let $\pi_{\theta}$ denote the victim VLA policy. Given a visual observation $x$ and a language instruction $\ell$, the policy predicts an action sequence $a=(a_1,\ldots,a_M)$. Let $a^\star=(a_1^\star,\ldots,a_M^\star)$ denote the attacker-specified target action sequence, and let $p \in [0,1]^{h\times w\times 3}$ denote the adversarial patch. Given a batch of clean scene frames $\{o_b\}_{b=1}^{B}$, we render the same patch into each frame through a compositing operator \begin{equation} \tilde{x}_b(p)=\mathcal{C}(o_b,p;m,T_b), \end{equation} where $m$ is the patch mask and $T_b$ is the geometric transform for the $b$-th frame. This captures the fact that the same physical patch may appear with different poses and placements across observations. We optimize the patch with the following targeted batch objective  \begin{equation} \mathcal{L}_{\mathrm{attack}}(p) = \frac{1}{B} \sum_{b=1}^{B} \sum_{i\in\mathcal{I}} w_i\, \mathcal{L}_{i} \left( y_{b,i}(p), a_i^\star \right), \label{eq:vla-loss} \end{equation} where $\mathcal{I}$ indexes the attacked action dimensions or tokens, $w_i$ are weighting coefficients, and $y_b(p)$ denotes the attack-visible output of the victim policy on the patched input $(\tilde{x}_b(p),\ell_b)$. The form of $y_b(p)$ depends on the setting: \begin{equation} y_b(p)= \begin{cases} \pi_{\theta}(\cdot \mid \tilde{x}_b(p),\ell_b), & \text{white-box},\\ \mathcal{Q}_{\theta}(\tilde{x}_b(p),\ell_b), & \text{black-box}, \end{cases} \end{equation} where the white-box output denotes differentiable action-token outputs such as logits, and $\mathcal{Q}_{\theta}$ denotes the black-box action-output interface that returns the predicted executable action sequence. Accordingly, we instantiate the per-dimension loss as \begin{equation} \mathcal{L}_{i}\left(y_{b,i}(p),a_i^\star\right) = \begin{cases} \mathrm{CE}\left(y_{b,i}(p),a_i^\star\right), & \text{white-box},\\ \left\| \bar{y}_{b,i}(p)-\bar{a}_i^\star \right\|_2^2, & \text{black-box}, \end{cases} \end{equation} where $\mathrm{CE}(\cdot,\cdot)$ is the targeted cross-entropy loss, and $\bar{y}_{b,i}(p)$ and $\bar{a}_i^\star$ denote predicted and target actions. For token-based policies under black-box access, $\mathcal{Q}_{\theta}$ returns the decoded continuous robot action before the MSE loss is evaluated.

\subsection{Diffusion-based Unrestricted Robotic Attacks}

As illustrated in Figure~\ref{fig:fig_main}, DURA formulates unrestricted patch generation as an attack-guided denoising problem. 
Instead of directly optimizing patch pixels, DURA starts from a benign seed patch and searches along the latent trajectory of a frozen pretrained diffusion model. 
This trajectory provides a natural generative space for patch content, while the targeted VLA loss steers the decoded patch toward the attacker-specified robot action. 
The same denoising trajectory is used in both white-box and black-box settings; only the way of estimating the per-step attack direction differs.

\noindent\textbf{Diffusion-guided Patch Optimization.}
We use a latent diffusion model~\citep{rombach2022high} with VAE encoder $\mathcal{E}$, VAE decoder $\mathcal{D}$, and frozen denoiser $\hat\epsilon_{\phi}$. 
Here, $\hat\epsilon_{\phi}(z_t,t)$ predicts the noise component of latent $z_t$ at diffusion timestep $t$. 
Given a benign seed patch crop $p_{\mathrm{clean}}$, DURA first encodes it into the latent space:
\begin{equation}
z_0^{\mathrm{clean}}=\mathcal{E}(p_{\mathrm{clean}}).
\end{equation}
Rather than starting from pure Gaussian noise, we partially noise this seed latent to an intermediate timestep $t_{\mathrm{start}}$, obtaining $z^{adv}_{t_{\mathrm{start}}}$ as the starting point of the adversarial denoising trajectory. 
This initialization preserves the natural structure of the seed patch while leaving sufficient latent-space freedom for the attack objective to reshape its content.

Starting from $z^{adv}_{t_{\mathrm{start}}}$, DURA alternates between denoising and adversarial steering. 
To keep the optimization close to a natural denoising path, we precompute a clean anchor trajectory $\{z_t^{\mathrm{anc}}\}$ by running an unperturbed DDIM pass~\citep{song2021denoising} from the same seed patch. 
At timestep $t$, the current adversarial latent is softly tied to the corresponding clean anchor:
\begin{equation}
z^{mix}_t=(1-\alpha_w)z^{adv}_t+\alpha_w z_t^{\mathrm{anc}},
\label{eq:anchor}
\end{equation}
where $\alpha_w\in[0,1]$ controls the anchor strength. 

DURA performs one DDIM denoising step and injects the attack update: 
\begin{equation}
u_{t-1}=\mathrm{DDIM}(z^{mix}_t,t;\hat\epsilon_{\phi}),
\qquad
z^{adv}_{t-1}=u_{t-1}-s\cdot g_t,
\label{eq:unified-update}
\end{equation}
where $s$ is the attack step size and $g_t$ is the latent attack direction. 
The DDIM step follows the pretrained diffusion prior, whereas the adversarial update moves the decoded patch toward the target robot action. 
The clean anchor regularizes this process by preventing the attack trajectory from drifting too far from the original denoising path.

\noindent\textbf{Attack Direction Estimation.}
The update rule in Eq.~\ref{eq:unified-update} reduces DURA to one central question: how to obtain the latent attack direction $g_t$ at each timestep. 
In the white-box setting, DURA decodes the current denoised latent, renders the candidate patch into the observation batch, and backpropagates the targeted VLA loss:
\begin{equation}
g_t^{\mathrm{WB}}
=
\nabla_{u_{t-1}}
\mathcal{L}_{\mathrm{attack}}
\left(\mathcal{D}(u_{t-1})\right).
\label{eq:whitebox-grad}
\end{equation}
 Here, $\mathcal{L}_{\mathrm{attack}}(\mathcal{D}(u_{t-1}))$ explicitly includes decoding, patch rendering, victim-policy evaluation, and the corresponding target-action loss computation. This estimates a local attack direction at the current denoising step, without backpropagating through the full DDIM trajectory.

In the black-box setting, gradients are unavailable and the attacker observes only action outputs.
DURA therefore estimates $g_t$ from query losses around the current denoised latent $u=u_{t-1}$.
To keep the search aligned with the diffusion prior, we sample candidates with the one-step forward noising distribution
$q_t(z_t\mid u)=\mathcal{N}(\sqrt{\alpha_t}u,(1-\alpha_t)I)$,
where $\alpha_t$ is the DDIM noise coefficient at timestep $t$.
Specifically, we sample $K$ perturbations as
$z_{t,k}=\sqrt{\alpha_t}u+\sqrt{1-\alpha_t}\epsilon_k$ with $\epsilon_k\sim\mathcal{N}(0,I)$,
decode the candidates into patches, render them into the same observation batch, and query the victim policy to obtain losses $\mathcal{L}_{\mathrm{attack}}^{(k)}$.
We employ a score-function estimator~\citep{williams1992simple}, the attack direction under black-box access is estimated as follows: 
\begin{equation}
g_t^{\mathrm{BB}}
=
\frac{1}{K}\sum_{k=1}^{K}
\left(\mathcal{L}^{(k)}_{\mathrm{attack}}-b\right)
\frac{\sqrt{\alpha_t}}{\sqrt{1-\alpha_t}}\,
\epsilon_k,
\label{eq:blackbox-grad}
\end{equation}
where $b$ is an optional variance-reduction baseline.
The full derivation of Eq.~\ref{eq:blackbox-grad} is provided in Appendix A.

\begin{algorithm}[t]
\caption{DURA Patch Optimization}
\label{alg:dura}
\begin{algorithmic}[1]
\REQUIRE Clean patch $p_{\mathrm{clean}}$, observation batch $\mathcal{B}$, target action $a^\star$, diffusion model $(\mathcal{E},\mathcal{D},\hat{\epsilon}_{\phi})$, access mode $\mathrm{mode}\in\{\mathrm{WB},\mathrm{BB}\}$
\ENSURE Adversarial patch $p_{\mathrm{adv}}$

\STATE $z^{\mathrm{clean}}_0 \gets \mathcal{E}(p_{\mathrm{clean}})$
\STATE Obtain $z^{\mathrm{adv}}_{t_{\mathrm{start}}}$ by partially noising $z^{\mathrm{clean}}_0$
\STATE Precompute clean anchor trajectory $\{z^{\mathrm{anc}}_t\}_{t=t_{\mathrm{start}}}^{0}$

\FOR{$t=t_{\mathrm{start}},\ldots,1$}
    \STATE ${z}^{mix}_t \gets (1-\alpha_w)z^{\mathrm{adv}}_t+\alpha_w z^{\mathrm{anc}}_t$
    \STATE $u_{t-1} \gets \mathrm{DDIM}({z}^{mix}_t,t;\hat{\epsilon}_{\phi})$
    \STATE $g_t \gets \textsc{EstimateDirection}(u_{t-1},\mathcal{B},a^\star,\mathrm{mode})$
    \STATE $z^{\mathrm{adv}}_{t-1} \gets u_{t-1}-s\cdot g_t$
\ENDFOR

\STATE $p_{\mathrm{adv}}\gets \mathcal{D}(z^{\mathrm{adv}}_0)$
\STATE \textbf{return} $p_{\mathrm{adv}}$
\end{algorithmic}
\end{algorithm}

\section{Experiments}
\label{sec:result}
 \begin{table*}[t]
  \centering

  \setlength{\tabcolsep}{4pt}
  \renewcommand{\arraystretch}{1.22}
  \resizebox{\textwidth}{!}{%
  \begin{tabular}{@{}lccccc!{\color{black!55}\vrule}ccccc@{}}
  \toprule
  & \multicolumn{5}{c}{\textbf{Simulated}} & \multicolumn{5}{c}{\textbf{Physical}} \\
  \cmidrule(lr){2-6} \cmidrule(lr){7-11}
  {Method} & {Spatial} & {Object} & {Goal} & {Long} & {\textbf{Avg}} & {Spatial} & {Object} & {Goal} & {Long} & {\textbf{Avg}} \\
  \midrule
  {\textsc{Benign} (no patch)} & 15.3 & 11.6 & 20.8 & 46.3 & 23.5 & 15.3 & 11.6 & 20.8 & 46.3 & 23.5 \\
  {\textsc{Clean patch}}        & 27.0 & 38.0 & 26.0 & 67.0 & 39.5 & 27.0 & 38.0 & 26.0 & 67.0 & 39.5 \\
  \midrule
  \multicolumn{11}{@{}l}{\textit{White-box attacks: label-supervised (requires action-token GT)}} \\
  \addlinespace[1pt]
  {\textsc{UMA}}   & \best{100.0} & \snd{98.8}   & \snd{99.0}   & \best{100.0} & \snd{99.5}   & \snd{96.6} & 56.4 & \snd{80.0} & \snd{82.0} & \snd{78.8} \\
  {\textsc{UPA}}   & \snd{96.2}         & 77.8         & 88.0        &\snd{96.8}         & 89.7         & 95.6 & \snd{57.0} & 57.2 & 72.8 & 70.7 \\
  {\textsc{UADA}}  & \best{100.0} & \best{100.0} & \best{100.0} & \best{100.0} & \best{100.0}
                   & \best{100.0} & \best{100.0} & \best{100.0} & \best{100.0} & \best{100.0} \\
  \midrule
  \multicolumn{11}{@{}l}{\textit{White-box attacks: target-only (no action-token GT)}} \\
  \addlinespace[1pt]
  {\textsc{TMA}}        & \best{100.0} & \best{100.0} & \best{100.0} & \snd{99.0}   & \snd{99.8}
                        & \snd{96.4} & 83.6 & \snd{74.4} & \snd{91.8} & \snd{86.6} \\
  {\textsc{FreezeVLA}}  & \snd{95.3}   & \snd{98.4}   & \snd{95.7}   & 92.2         & 95.4
                        & {\na} & {\na} & {\na} & {\na} & {\na} \\
  % \rowcolor{ourrow}
  {\ours}               & \best{100.0\,\gain{4.7}} & \best{100.0\,\gain{1.6}} & \best{100.0\,\gain{4.3}} & \best{100.0\,\gain{1.0}} & \best{100.0\,\gain{0.2}}
                        & \best{100.0\,\gain{3.6}} & \best{100.0\,\gain{16.4}} & \best{100.0\,\gain{25.6}} & \best{100.0\,\gain{8.2}} & \best{100.0\,\gain{13.4}} \\
  \midrule
  \multicolumn{11}{@{}l}{\textit{Black-box attacks: target-only}} \\
  \addlinespace[1pt]
  {\textsc{TMA-NES}}    & \snd{43.0} & \snd{36.0} & \snd{30.0} & \snd{60.0} & \snd{42.3}
                        & \snd{41.0} & \snd{35.0} & \snd{28.0} & \snd{53.0} & \snd{39.3} \\
  % \rowcolor{ourrow}
  {\ours}               & \best{100.0\,\gain{57.0}} & \best{96.0\,\gain{60.0}} & \best{68.0\,\gain{38.0}} & \best{80.0\,\gain{20.0}} & \best{86.0\,\gain{43.7}}
                        & \best{97.0\,\gain{56.0}} & \best{70.0\,\gain{35.0}} & \best{76.0\,\gain{48.0}} & \best{74.0\,\gain{21.0}} & \best{79.3\,\gain{40.0}} \\
  \bottomrule
  \end{tabular}}
    \caption{\textbf{Attack Success Rate (ASR, \%, $\uparrow$) on LIBERO} for
    OpenVLA-7B across four task suites. Methods are grouped by supervision: \emph{label-supervised}
    attacks use ground-truth action tokens, while \emph{target-only} attacks need only a fixed
    target. \textbf{Bold} and \underline{underline} mark the best and second-best within each group;
    $\uparrow$ is the gain of \textsc{Ours} over the next-best distinct score.}
    \label{tab:main_results}
  \end{table*}
\subsection{Experimental Setup}
\begin{figure*}[t] 
    \centering 
    \includegraphics[width=1.0\textwidth]{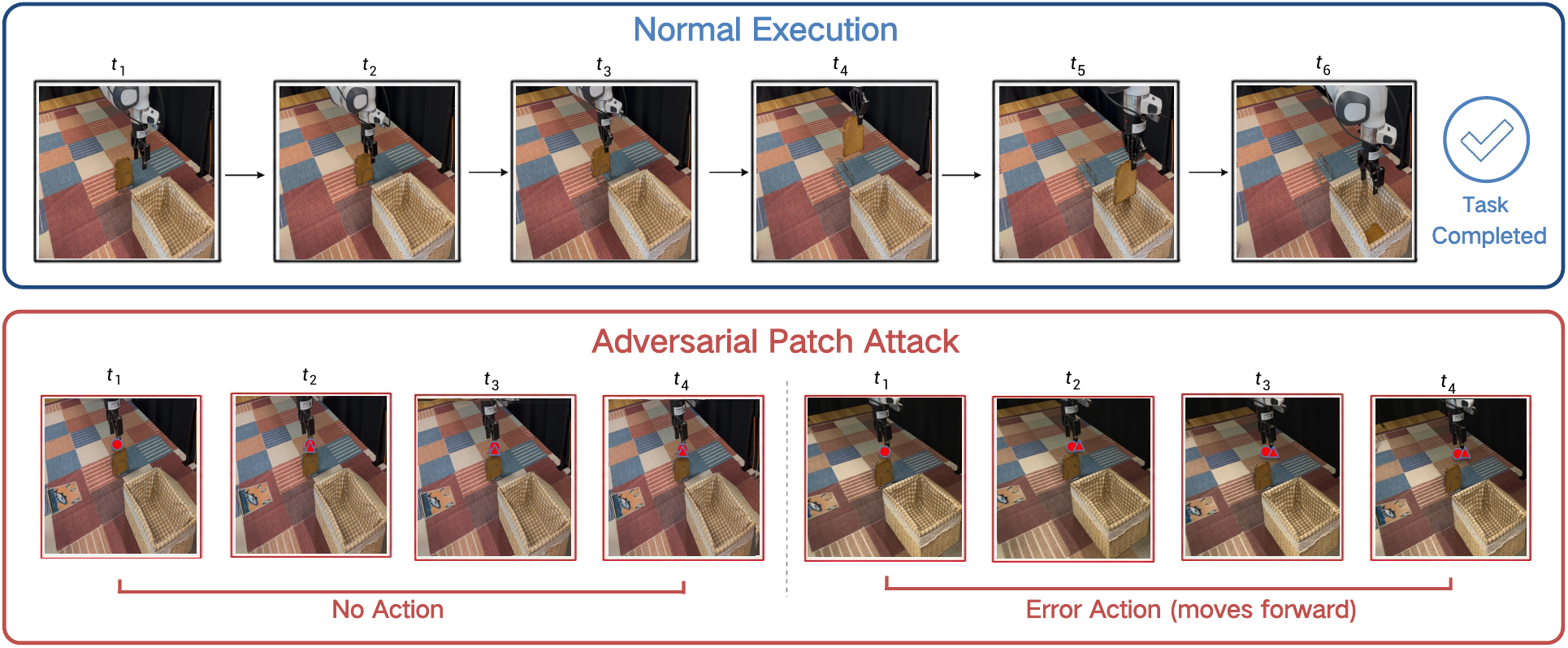}
    \caption{\textbf{Real-robot targeted attack on a Franka arm} (task: place the bread slice into the basket). (a) Without the patch, the arm completes the task. (b) Inserting the printed patch into the camera view drives the arm to the target action and it stays still; removing it lets the arm resume; inserting it again drives the arm to the target action once more. Here, $\triangle$ denotes the position in the previous frame, while $\circ$ denotes the position in the current frame. The behavior tracks the presence of the patch, which shows the attack is controllable and repeatable on real hardware.}
    \label{fig:realrobot}
\end{figure*}
\noindent\textbf{Datasets and Models.}
We evaluate our method on two widely used robotic manipulation benchmarks, LIBERO~\citep{liu2023libero} and BridgeData~V2~\citep{walke2023bridgedata}. To cover different VLA action representations, we consider two representative models, OpenVLA~\citep{kim2024openvla} and $\pi_0$-FAST~\citep{karl2025fast}. For OpenVLA, we use checkpoints fine-tuned on the four LIBERO suites. For $\pi_0$-FAST, we use its LIBERO fine-tuned checkpoint.

\noindent\textbf{Attack Configuration.}
We study targeted attacks in both white-box and black-box settings. The patch is optimized over batches of frames sampled from multiple rollouts, so that it generalizes across scenes, robot states, and instructions rather than overfitting to a single observation. We compare against strong white-box baselines, including UMA, UPA, UADA, TMA, and FreezeVLA, and against query-based black-box baselines. Since there is no directly comparable query-only baseline for OpenVLA, we adapt TMA into a black-box NES variant, denoted TMA-NES~\cite{ilyas2018black,chen2017zoo}. For $\pi_0$-FAST, we additionally report UPA-RFAS.

\noindent\textbf{Patch Generation and Evaluation Setup.}
 We craft adversarial patches under three generating settings. The \emph{simulation setting} uses the Openvla-7B model trained in simulation on the LIBERO ~\citep{kim2024openvla}, while the \emph{physical setting} uses a model trained on real-world data from BridgeData V2 with the openvla-7B model~\citep{walke2023bridgedata}.
We then evaluate the generated patches on victim models trained on different task suites, which differ in both data source and task objective, to rigorously verify the robustness and effectiveness of our method. For the \emph{real-world evaluation}, we deploy the attack on a Franka robot arm, a 7-DoF manipulator equipped with a parallel-jaw gripper and observed by a single fixed RGB camera. We print the optimized adversarial patch and place it in the robot workspace within the camera view, while keeping the task instruction and policy weights unchanged so that the only intervention is the presence of the printed patch. This setup tests whether the optimized patch can induce the intended target action under real imaging conditions, including printing artifacts, lighting variations, camera perspective, and placement noise.

\begin{table}[t]
    \centering

    {\small
    \setlength{\tabcolsep}{6.20pt}
    \renewcommand{\arraystretch}{1.10}

    \begin{tabular}{@{}lccccc@{}}
    \toprule
    Method & Spatial & Object & Goal & Long & \textbf{Avg} \\
    \midrule
    \textsc{Benign} (no patch)
        & 12.0 & 9.0 & 6.0 & 31.0 & 14.5 \\
    \textsc{Clean patch}
        & 21.8 & 9.2 & 10.8 & 41.4 & 20.8 \\
    \midrule

    \multicolumn{6}{@{}l}{\textit{White-box attacks: target-only}} \\
    \addlinespace[1pt]

    \textsc{TMA}
        & \snd{55.0}
        & \snd{46.0}
        & \snd{78.0}
        & \snd{53.0}
        & \snd{58.0} \\

    \ours
        & \best{100.0}
        & \best{100.0}
        & \best{100.0}
        & \best{100.0}
        & \best{100.0} \\
    \midrule

    \multicolumn{6}{@{}l}{\textit{Black-box attacks: target-only}} \\
    \addlinespace[1pt]

    \textsc{TMA-NES}
        & \snd{31.0}
        & \snd{14.0}
        & \snd{34.0}
        & \snd{46.1}
        & \snd{31.3} \\

    \textsc{UPA-RFAS}
        & 27.0
        & \snd{14.0}
        & 21.0
        & 42.0
        & 26.0 \\

    \ours
        & \best{100.0}
        & \best{100.0}
        & \best{100.0}
        & \best{100.0}
        & \best{100.0} \\
    \bottomrule
    \end{tabular}
    }

    \caption{Attack Success Rate (ASR, \%, $\uparrow$) on LIBERO
    for $\boldsymbol{\pi_0}$-FAST across four task suites.
    \emph{Target-only} attacks require only a fixed target action.
    Bold and underlined values denote the best and second-best
    results within each group, respectively.}
    \label{tab:pi0fast_results}
\end{table}

\noindent\textbf{Implementation Details.}
Our attack optimizes the patch in the VAE latent space of a frozen diffusion model~\citep{rombach2022high}. A clean seed patch is noised to $t_{\mathrm{start}}=0.5$ and optimized for 200 DDIM steps~\citep{song2021denoising}, with anchor weight $\alpha_w=0.2$. The target-action loss uses weights  $1.0$, $0.5$, and $0.2$ for translation, rotation, and gripper dimensions, respectively. In the black-box setting, $K=2048$ denotes the number of queries per optimization update. For fair comparison, both DURA and TMA-NES use 100 optimization updates, leading to a default budget of $100\times K$ queries per patch. 
% In the black-box setting, each update uses $K=2048$ antithetic Gaussian queries, and each perturbed latent is mapped back to a clean estimate via Tweedie's formula before decoding.

\noindent\textbf{Evaluation Metrics.}
% We report attack success rate (ASR) and attack precision (AP). 
% The Attack Success Rate (ASR) measures the fraction of evaluation rollouts in which the attacked policy fails the task. To distinguish targeted attacks from incidental performance degradation, the Attack Precision (AP) further measures how often the executed behavior matches the attacker-specified target action. Unless otherwise stated, we evaluate each LIBERO suite over its 10 tasks with 50 rollouts each, yielding 500 rollouts per suite.
Attack Success Rate (ASR) is computed at the rollout level and reports the fraction of rollouts in which the attacked policy fails to complete the task:

  \begin{equation}
  \mathrm{ASR} = \frac{N_{\text{fail}}}{N},
  \end{equation}
  where $N$ denotes the total number of evaluation rollouts and $N_{\text{fail}}$ the number of failed rollouts. A higher ASR indicates a
  stronger disruption of task execution.

  Attack Precision (AP) is computed at the action-step level and quantifies the consistency between the executed behavior and the
  attacker-specified target action. For the $i$-th trajectory, let $T^{(i)}$ denote its total number of action steps and
  $T^{(i)}_{\text{target}}$ the number of steps matching the target action (e.g., \emph{no-action} or \emph{move-forward}). AP is defined as
  the per-trajectory ratio averaged over all $N$ trajectories:

  \begin{equation}
  \mathrm{AP} = \frac{1}{N} \sum_{i=1}^{N} \frac{T^{(i)}_{\text{target}}}{T^{(i)}}.
  \end{equation}
  A higher AP indicates that the policy is steered more consistently toward the target action, which separates targeted control from incidental performance degradation.

\subsection{Main Results}
  \label{sec:main}

% \noindent\textbf{Simulation Results.}
% Table~\ref{tab:main_results} reports ASR on the four LIBERO suites for OpenVLA,
% under both white-box and black-box access and with patches crafted in the
% simulation and physical settings. Under white-box access, DURA reaches $100\%$
% ASR on every suite in both settings. DURA is a target-only attack and uses no
% ground-truth action tokens. It still matches \textsc{UADA}, the only
% label-supervised baseline that also reaches $100\%$, and surpasses the strongest
% target-only baselines, \textsc{TMA} and \textsc{FreezeVLA} ($99.8\%$ and $95.4\%$
% average in simulation). The same patch trained on real-world Bridge~V2 frames again
% reaches $100\%$ average, while \textsc{TMA} drops to $86.6\%$, which shows that
% DURA does not rely on simulated textures.

% The black-box setting is where DURA gains the most. Using queries alone, it
% reaches $86.0\%$ average ASR in simulation and $79.3\%$ in the physical setting,
% against $42.2\%$ and $39.2\%$ for \textsc{TMA-NES}, the only black-box baseline
% that produces a meaningful attack. The margin is $43.8$ and about $40$ points, and
% per-suite ASR is as high as $100\%$ on Spatial. These results show that a strong
% attack on VLA models does not require white-box access. Across all blocks, the
% benign failure rate is $23.5\%$ and an unoptimized clean patch reaches only
% $39.5\%$, so the gains come from the optimized patch rather than the sticker itself.

\noindent\textbf{Quantitative Results.}
Table~\ref{tab:main_results} summarizes the OpenVLA results on LIBERO. Under white-box access, DURA achieves 100\% ASR on all suites in both simulated and physical patch settings. Although DURA is target-only and does not use ground-truth action tokens, it matches the strongest label-supervised baseline UADA and outperforms prior target-only attacks. Under black-box access, DURA achieves 86.0\% ASR with simulated patches and 79.3\% with physical patches, improving over TMA-NES by 43.7 and 40.0 points, respectively. In contrast, the benign policy and clean patch obtain only 23.5\% and 39.5\% ASR, confirming that the failures are caused by the optimized adversarial patch rather than patch insertion alone. Beyond task failure, DURA also attains a high attack precision (AP), showing that the failures stem from the intended target action rather than incidental
out-of-distribution degradation. Because AP is only well-defined for target-only attacks, we report applicable AP results separately in Appendix C.

% \noindent\textbf{Real-Robot Results.}
%   An ideal attack must succeed not only in simulation but on a real robot, since the
%   physical-world threat is what ultimately matters. We test this property directly. We
%   run a Franka arm on a pick-and-place task, where the arm places a bread slice into a
%   basket, print the optimized patch, and bring it into the camera view during execution.
%   Figure~\ref{fig:realrobot} compares a normal rollout with an attacked one.

%   DURA succeeds on the real arm, and it does so in a controllable way across three
%   stages. While the arm is executing the task, we bring the patch into view, and the arm
%   switches to the target action and stays still. We then remove the patch, and the arm
%   resumes and finishes the task, which shows that the effect comes from the patch and that
%   the attack is precise. We bring the patch back into view, and the arm returns to the
%   target action once more. The target action appears and disappears with the patch, so
%   DURA gives an attacker repeatable control over a physical robot. This result confirms
%   that DURA meets the real-world deployability that an ideal attack requires.
\begin{figure}[t]
    \centering
    \includegraphics[width=1.0\columnwidth]{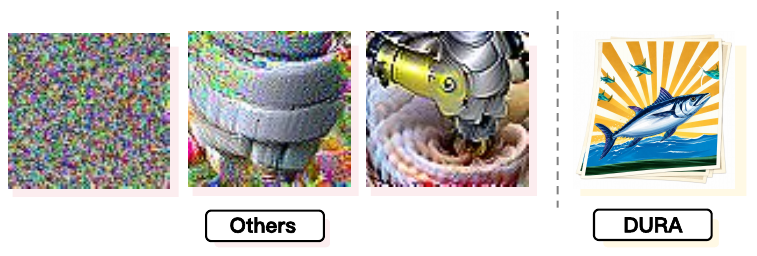}\\
   \caption{\textbf{Qualitative comparison of adversarial patch naturalness.}
From left to right, we show patches generated by UADA, UPA, TMA, and DURA under the main experimental configuration. Compared with the noise-like or visually irregular patterns produced by the baselines, the DURA patch exhibits a more coherent appearance and recognizable semantic content.}
    \label{fig:patch_examples}
\end{figure}
\begin{figure}[h]
    \centering
    \includegraphics[width=1.0\columnwidth]{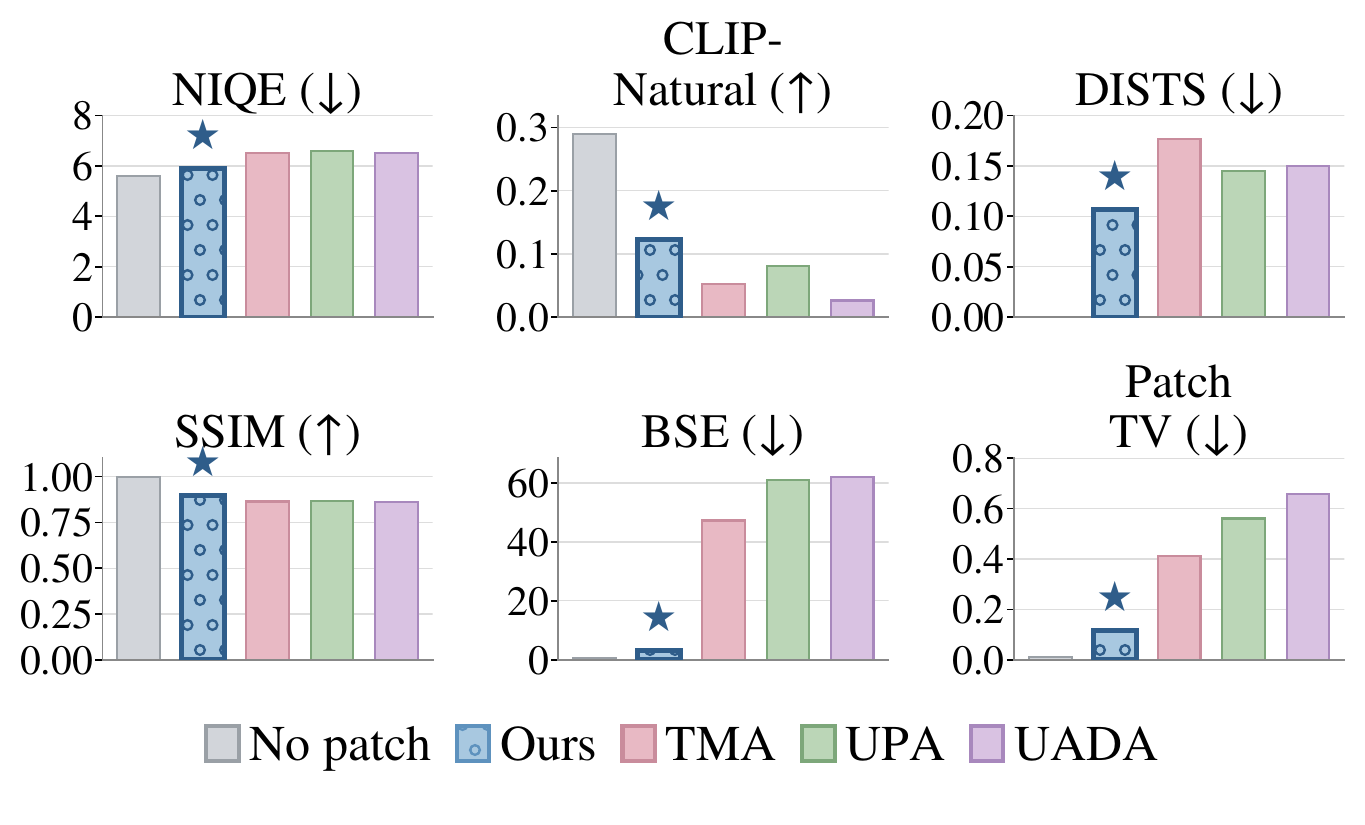}\\
   \caption{\textbf{Visual naturalness comparison.}
The figure reports six visual-quality metrics for DURA($\star$), \textsc{TMA}, \textsc{UPA}, and \textsc{UADA}, together with the no-patch clean reference baseline. DURA stays closest to the no-patch baseline among attack methods, indicating better local naturalness and fewer visible artifacts.}
    \label{fig:Visual naturalness}
\end{figure}
\subsection{Additional Analyses}
\label{sec:diag}
\noindent\textbf{Real-Robot Results.}
We evaluate DURA on a physical Franka arm in a bread-in-basket task. As shown in Figure~\ref{fig:realrobot}, the clean rollout follows the intended pick-and-place trajectory and completes the task. In contrast, when the printed adversarial patch appears in the camera view, the robot deviates from normal execution and enters attacker-specified behaviors, including freezing action~\citep{wang2025freezevla} and moving forward erroneously~\citep{wang2025exploring}. 
With the policy, instruction, and robot setup unchanged, DURA transfers to real hardware and induces diverse physical failure modes beyond generic task failure.

\noindent\textbf{Visual Naturalness.} Figure~\ref{fig:Visual naturalness}
evaluates the visual quality of the optimized patches using six metrics:
NIQE~\citep{mittal2012making},
CLIP-Natural~\citep{radford2021learning,wang2023CLIPIQA},
DISTS~\citep{ding2020image},
SSIM~\citep{wang2004image},
Boundary Seam Energy (BSE)~\citep{taeg2010tpt},
and Patch Total Variation (Patch TV)~\citep{rudin1992nonlinear},
with the evaluation protocol deferred to Appendix D. Overall, DURA is visually closer to the no-patch scene than the pixel-space baselines. This trend appears not only in perceptual similarity metrics, but also in local artifact measurements around the patch region. In particular, the pixel-space attacks tend to introduce high-frequency textures and visible boundary seams, whereas the diffusion-guided patch remains smoother and more consistent with the surrounding scene. A qualitative comparison under the main experimental configuration is shown in Figure~\ref{fig:patch_examples}. These results support that DURA improves the stealthiness of the patch through the diffusion prior, instead of relying on conspicuous adversarial patterns.\looseness=-1

% \noindent\textbf{Query budget.}

% The black-box attack estimates each update from $K$ queries, so we study how ASR
% depends on $K$ on LIBERO with OpenVLA. Figure~\ref{fig:Query budget K} shows that
% DURA improves steadily with the budget, from $42\%$ ASR at $K{=}128$ to $100\%$ at
% $K{=}2048$. The \textsc{TMA-NES} baseline stays between $40\%$ and $48\%$ regardless of
% $K$. This shows that our diffusion-guided estimator turns extra queries into a stronger
% attack, while patch-space search does not.
% \begin{figure}[h]
%     \centering
%     \includegraphics[width=0.95\columnwidth]{Figures/openvla_spatial_k_ablation_asr_comparison.png}\\
    
%     \caption{\textbf{Query budget K.} Black-box ASR versus query
%     budget $K$ on LIBERO; DURA scales with $K$ while \textsc{TMA-NES} stays flat.}
%     \label{fig:Query budget K}
% \end{figure}
\begin{figure}[t]
    \centering
    \includegraphics[width=0.95\columnwidth]{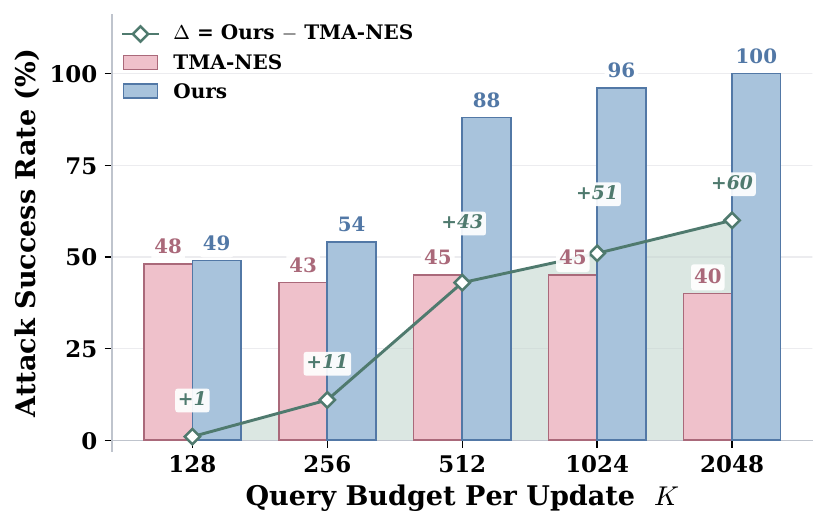}\\
    
    \caption{\textbf{Query budget K.} Black-box ASR versus query
    budget $K$ on LIBERO. DURA already achieves a high ASR at $K{=}512$ and improves further as the budget increases, whereas \textsc{TMA-NES} remains at a consistently low ASR.}
    \label{fig:Query budget K}
\end{figure}

\noindent\textbf{Query Budget.}
Figure~\ref{fig:Query budget K} studies the effect of the black-box query budget. DURA and TMA-NES use the same batch of frames and the same number of optimization iterations. For each update, both methods use $K$ black-box action-output queries, and we sweep the per-update budget from 128 to 2048 while matching the total queries between methods. DURA improves steadily as $K$ increases, with ASR rising from 49\% at $K{=}128$ to 100\% at $K{=}2048$. Notably, DURA already achieves a relatively high ASR at $K{=}512$, demonstrating effective attacks with a moderate query budget. In contrast, TMA-NES remains around 40\%--48\% under the same budgets. This gap indicates that DURA uses the same query budget more effectively by performing black-box search in a diffusion-guided latent space. Importantly, practical attacks need not use the query budget required for 100\% ASR, since DURA already achieves a high ASR with substantially fewer queries, enabling a flexible trade-off between attack effectiveness and query cost.

% \noindent\textbf{Patch size.}

% We vary the patch side length from 0.2\% to 10\% pixels and report the mean ASR over
% three seeds. Figure~\ref{fig:Patch size}(b) shows a clear threshold: patches up to $30$
% pixels reach only $21$ to $30\%$ ASR, sizes of $40$ to $50$ pixels give high but unstable
% ASR, and sizes of $60$ pixels and above saturate near $100\%$. 
% \begin{figure}[t]
% \centering
% \includegraphics[width=0.95\columnwidth]{Figures/openvla_spatial_patch_size_asr.png}\\
% {\small Patch size}
% \caption{\textbf{Patch size.}
% White-box ASR versus patch size on LIBERO-Spatial; ASR rises past a size threshold
% and saturates.}
% \label{fig:Patch size}
% \end{figure}

% \noindent\textbf{Impact of Patch Size.}
% We further study how the patch size and placement affect the attack, and find that DURA stays effective even with a small-area patch; the full analysis is provided in Appendix~\ref{app:patch_size}.

% \noindent\textbf{Robustness across victim models.}
% So far we have attacked OpenVLA. To test whether DURA depends on a particular
% architecture, we apply it to Pi0-FAST, a flow-matching VLA with a different action
% representation. Table~\ref{tab:pi0fast_results} reports ASR on the four LIBERO suites.
% DURA reaches $100\%$ ASR on every suite, in both the white-box and black-box settings.
% The strongest baselines stay far below, with \textsc{TMA} at $58.0\%$ white-box and
% \textsc{TMA-NES} at $26.2\%$ black-box on average. This shows that the attack carries
% over to a different model family rather than exploiting one design.
\noindent\textbf{Impact of the Victim Model.}
We further evaluate DURA on $\pi_0$-FAST model to examine whether the attack transfers across VLA models and action architectures. Unlike OpenVLA, which predicts discrete action tokens, $\pi_0$-FAST leverages action chunks. As shown in Table~\ref{tab:pi0fast_results}, DURA achieves 100\% ASR on all four LIBERO task suites in both white-box and black-box settings. In comparison, the strongest baselines reach only 58.0\% ASR in the white-box setting and 31.3\% in the black-box setting. These results indicate that DURA can reliably induce target behaviors and achieve successful attacks across different VLA action formulations.

% \noindent\textbf{Robustness to Input Transformations.}
% We also test DURA against common input-transformation defenses (JPEG compression,bit-depth reduction, and Gaussian noise), evaluating each with both ASR and AP\citep{guo2017countering,xu2017feature}. DURA retains a high ASR and an AP close to the no-defense level under all defenses. Full results are in Appendix~\ref{app:defense}.

\begin{figure}[t]
    \centering
    \includegraphics[width=1.0\columnwidth]{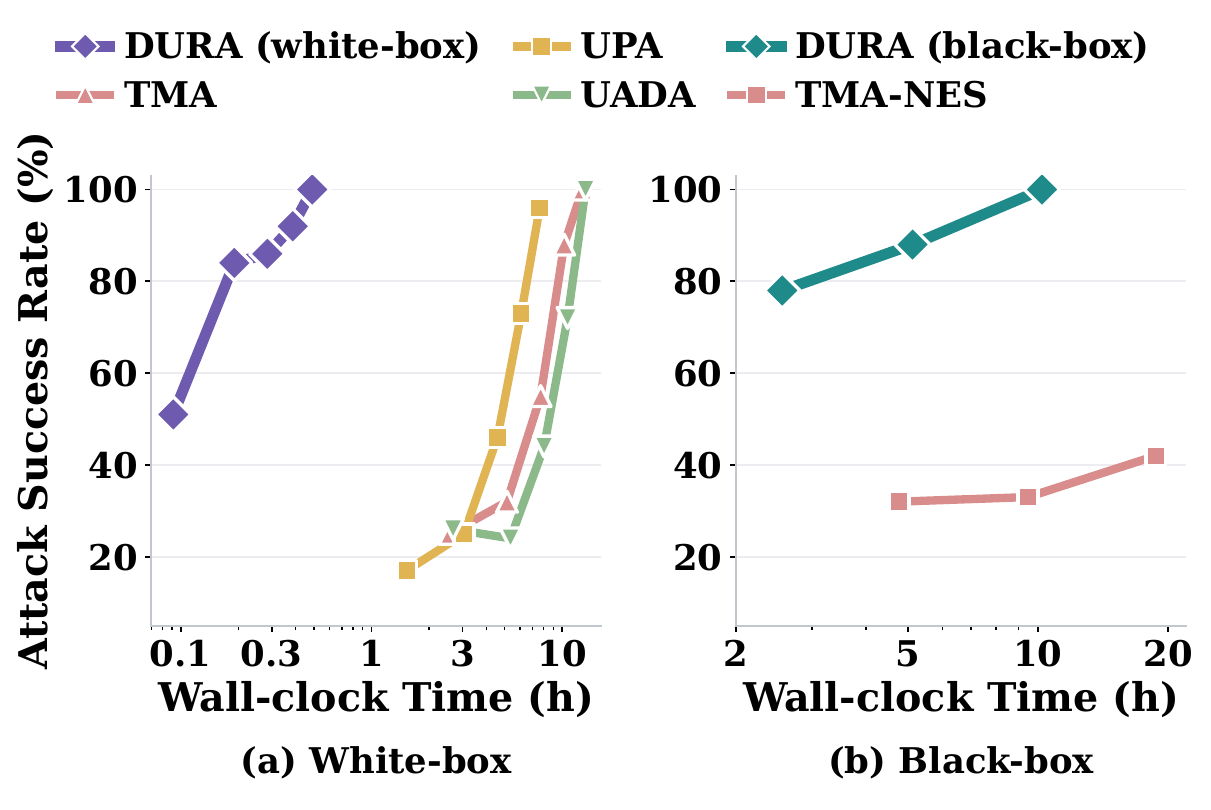}\\
\caption{\textbf{Time efficiency.} Time--ASR trade-off on OpenVLA under (a) white-box and (b) black-box access. The x-axis reports single-GPU wall-clock time in hours. DURA reaches high attack success substantially earlier than direct pixel-space optimization baselines in both settings.} 
    \label{fig:Time efficiency}
\end{figure}
% \noindent\textbf{Time Efficiency Analysis} 
% As shown in Figure~\ref{fig:Time efficiency}, our method is evaluated against four representative baselines, TMA, UPA, and UADA under the white-box setting and TMA-NES under the black-box setting, whose attacks rely on lengthy iterative optimization with a fixed per-step query budget.we keep the victim model, the evaluation suite, and the patch configuration fixed, and vary only the wall-clock time budget. Our method reaches a high success rate much earlier in the optimization, achieving about 80\% attack success apperate at $1\times$ and rising to 100\% within $3\times$ of that budget. On the white-box setting in particular, it attains comparable success $20$ to $64\times$ faster than TMA, UPA, and UADA, outperforming the strongest baseline by more than an order of magnitude in time. This efficiency advantage consistently holds across both threat models, where in the black-box setting our method reaches 96\% success while TMA-NES plateaus near 42\% even at roughly $7\times$ the time.
\noindent\textbf{Time Efficiency.}
Figure~\ref{fig:Time efficiency} compares the time--ASR trade-off on OpenVLA. All runtimes are measured as single-GPU wall-clock time on an H200 with batch size 16, while the victim model, evaluation suite, and patch setting are kept fixed. For the direct pixel-space baselines, we trace the curves by varying the number of optimization steps; for DURA, we vary the number of DDIM steps. In the white-box setting, DURA reaches high ASR within minutes, whereas TMA, UPA, and UADA require hours to approach comparable success rates.  In the black-box setting, DURA achieves higher ASR with longer optimization time, whereas TMA-NES remains less effective even with extended runtimes. Overall, DURA provides a more favorable trade-off between runtime and ASR under both access settings.\looseness=-1

% \section{Conclusion}                                                                                                          
% \label{sec:conclusion}
% In this work, DURA shows that VLA models can be attacked
%   through nothing more than the model's inference interface. This means the attack can be
%   carried out against real-world deployed systems, not just academic                                                                           
%   benchmarks. VLA models directly drive robots in the physical world, so
%   a real-world deployable attack on them no longer causes only digital                                                                         
%   errors but actual physical damage. DURA makes it concrete that this
%   risk is no longer hypothetical, and any robot running a VLA model is                                                                         
%   now a potential target rather than a future one. Therefore, VLA safety
%   can no longer be treated as a side topic. The research community needs                                                                       
%   to invest in protecting these models with the same seriousness it has                                                                        
%   invested in advancing their capability.
\section{Conclusion}

We presented DURA, a diffusion-guided unrestricted adversarial patch attack for VLA models. By optimizing along the latent trajectory of a pretrained diffusion model, DURA generates localized, natural-looking patches that steer robot policies toward attacker-specified actions under both white-box and action-output black-box access. Experiments on OpenVLA and $\pi_0$-FAST across LIBERO show that DURA improves attack success, target-action control, visual naturalness, and efficiency over pixel-space and query-based baselines. Printed-patch Franka experiments further suggest that the attack can persist under physical execution conditions.

These findings expose a risk for embodied AI: adversarial control can be embedded in visually ordinary workspace content, not only in synthetic or noise-like perturbations. Once placed in the robot's camera view, such content may turn the physical workspace itself into an attack surface. This calls for VLA defenses that evaluate natural-looking, physically placeable adversarial content, rather than assuming attacks will appear as conspicuous artifacts.
% \section*{Acknowledgments}
% AAAI is especially grateful to Peter Patel Schneider for his work in implementing the original aaai.sty file, liberally using the ideas of other style hackers, including Barbara Beeton. We also acknowledge with thanks the work of George Ferguson for his guide to using the style and BibTeX files --- which has been incorporated into this document --- and Hans Guesgen, who provided several timely modifications, as well as the many others who have, from time to time, sent in suggestions on improvements to the AAAI style. We are especially grateful to Francisco Cruz, Marc Pujol-Gonzalez, and Mico Loretan for the improvements to the Bib\TeX{} and \LaTeX{} files made in 2020.

% The preparation of the \LaTeX{} and Bib\TeX{} files that implement these instructions was supported by Schlumberger Palo Alto Research, AT\&T Bell Laboratories, Morgan Kaufmann Publishers, The Live Oak Press, LLC, and AAAI Press. Bibliography style changes were added by Sunil Issar. \verb+\+pubnote was added by J. Scott Penberthy. George Ferguson added support for printing the AAAI copyright slug. Additional changes to aaai2027.sty and aaai2027.bst have been made by Francisco Cruz, Marc Pujol-Gonzalez, and Mico Loretan.

% \bigskip
% \noindent Thank you for reading these instructions carefully. We look forward to receiving your electronic files!

\clearpage
\bibliography{aaai2027}
\clearpage
% \begin{figure}[t]
% \centering
% \includegraphics[width=0.95\columnwidth]{Figures/openvla_spatial_patch_size_asr.png}\\
% {\small Patch size}
% \caption{\textbf{Patch size.}
% White-box ASR versus patch size on LIBERO-Spatial; ASR rises past a size threshold
% and saturates.}
% \label{fig:Patch size}
% \end{figure}

\appendix
\setcounter{secnumdepth}{2}
\section{Derivation of the Black-box Attack Direction}
\label{app:bb_derivation}

At DDIM timestep $t$, let $u=u_{t-1}$ denote the latent after one DDIM denoising step and before the adversarial update. 
To estimate the attack direction under black-box access, DURA samples one-step noised query latents around $u$:
\begin{equation}
q_t(z_t\mid u)
=
\mathcal{N}
\left(
\sqrt{\alpha_t}u,
(1-\alpha_t)I
\right),
\label{eq:app-query-dist}
\end{equation}
or equivalently,
\begin{equation}
z_t
=
\sqrt{\alpha_t}u
+
\sqrt{1-\alpha_t}\epsilon,
\qquad
\epsilon\sim\mathcal{N}(0,I).
\label{eq:app-z-sample}
\end{equation}
For each sampled $z_t$, the frozen diffusion and decoding pipeline constructs a candidate patch, which is rendered into the observation batch and evaluated through the victim policy. 
We denote this black-box query pipeline by $\Phi_t(z_t)$.

We define the smoothed attack objective as
\begin{equation}
J_t(u)
=
\mathbb{E}_{z_t\sim q_t(\cdot\mid u)}
\left[
\mathcal{L}_{\mathrm{attack}}
\left(
\Phi_t(z_t)
\right)
\right].
\label{eq:app-smoothed-objective}
\end{equation}
Using the score-function identity, its gradient with respect to $u$ is
\begin{align}
\nabla_u J_t(u)
&=
\nabla_u
\int
\mathcal{L}_{\mathrm{attack}}
\left(
\Phi_t(z_t)
\right)
q_t(z_t\mid u)
\,dz_t
\\
&=
\int
\mathcal{L}_{\mathrm{attack}}
\left(
\Phi_t(z_t)
\right)
\nabla_u q_t(z_t\mid u)
\,dz_t
\\
&=
\mathbb{E}_{z_t\sim q_t(\cdot\mid u)}
\left[
\mathcal{L}_{\mathrm{attack}}
\left(
\Phi_t(z_t)
\right)
\nabla_u\log q_t(z_t\mid u)
\right].
\label{eq:app-score-identity}
\end{align}

It remains to compute the score term. 
From Eq.~\ref{eq:app-query-dist},
\begin{equation}
\log q_t(z_t\mid u)
=
C
-
\frac{1}{2(1-\alpha_t)}
\left\|
z_t-\sqrt{\alpha_t}u
\right\|_2^2,
\end{equation}
where $C$ is independent of $u$. 
Thus,
\begin{align}
\nabla_u\log q_t(z_t\mid u)
&=
\frac{\sqrt{\alpha_t}}{1-\alpha_t}
\left(
z_t-\sqrt{\alpha_t}u
\right).
\label{eq:app-score-raw}
\end{align}
Substituting Eq.~\ref{eq:app-z-sample} into Eq.~\ref{eq:app-score-raw} gives
\begin{equation}
\nabla_u\log q_t(z_t\mid u)
=
\frac{\sqrt{\alpha_t}}{\sqrt{1-\alpha_t}}
\epsilon.
\label{eq:app-score-term}
\end{equation}

Combining Eq.~\ref{eq:app-score-identity} and Eq.~\ref{eq:app-score-term}, we obtain
\begin{equation}
\nabla_u J_t(u)
=
\mathbb{E}_{\epsilon\sim\mathcal{N}(0,I)}
\left[
\mathcal{L}_{\mathrm{attack}}
\left(
\Phi_t(z_t)
\right)
\frac{\sqrt{\alpha_t}}{\sqrt{1-\alpha_t}}
\epsilon
\right].
\label{eq:app-gradient-expectation}
\end{equation}

With $K$ Monte Carlo samples, DURA draws
\begin{equation}
z_{t,k}
=
\sqrt{\alpha_t}u
+
\sqrt{1-\alpha_t}\epsilon_k,
\qquad
\epsilon_k\sim\mathcal{N}(0,I),
\end{equation}
and queries the corresponding losses
\begin{equation}
\mathcal{L}_{\mathrm{attack}}^{(k)}
=
\mathcal{L}_{\mathrm{attack}}
\left(
\Phi_t(z_{t,k})
\right).
\end{equation}
The resulting estimator is
\begin{equation}
g_t^{\mathrm{BB}}
=
\frac{1}{K}
\sum_{k=1}^{K}
\left(
\mathcal{L}_{\mathrm{attack}}^{(k)}-b
\right)
\frac{\sqrt{\alpha_t}}{\sqrt{1-\alpha_t}}
\epsilon_k,
\label{eq:app-bb-estimator}
\end{equation}
where $b$ is a variance-reduction baseline. 
This baseline does not change the expectation, since
\begin{align}
\mathbb{E}_{z_t\sim q_t(\cdot\mid u)}
\left[
\nabla_u\log q_t(z_t\mid u)
\right]
&=
\int
\nabla_u q_t(z_t\mid u)
\,dz_t
\\
&=
\nabla_u
\int
q_t(z_t\mid u)
\,dz_t
=
0.
\end{align}
Therefore,
\begin{equation}
\mathbb{E}
\left[
g_t^{\mathrm{BB}}
\right]
=
\nabla_u J_t(u).
\end{equation}

DURA then applies this estimate in the shared update rule:
\begin{equation}
z^{adv}_{t-1}
=
u_{t-1}
-
s\cdot g_t^{\mathrm{BB}}.
\end{equation}

\section{Impact of Patch Size}\label{app:patch_size}

\begin{figure}[h]
\centering
\includegraphics[width=0.95\columnwidth]{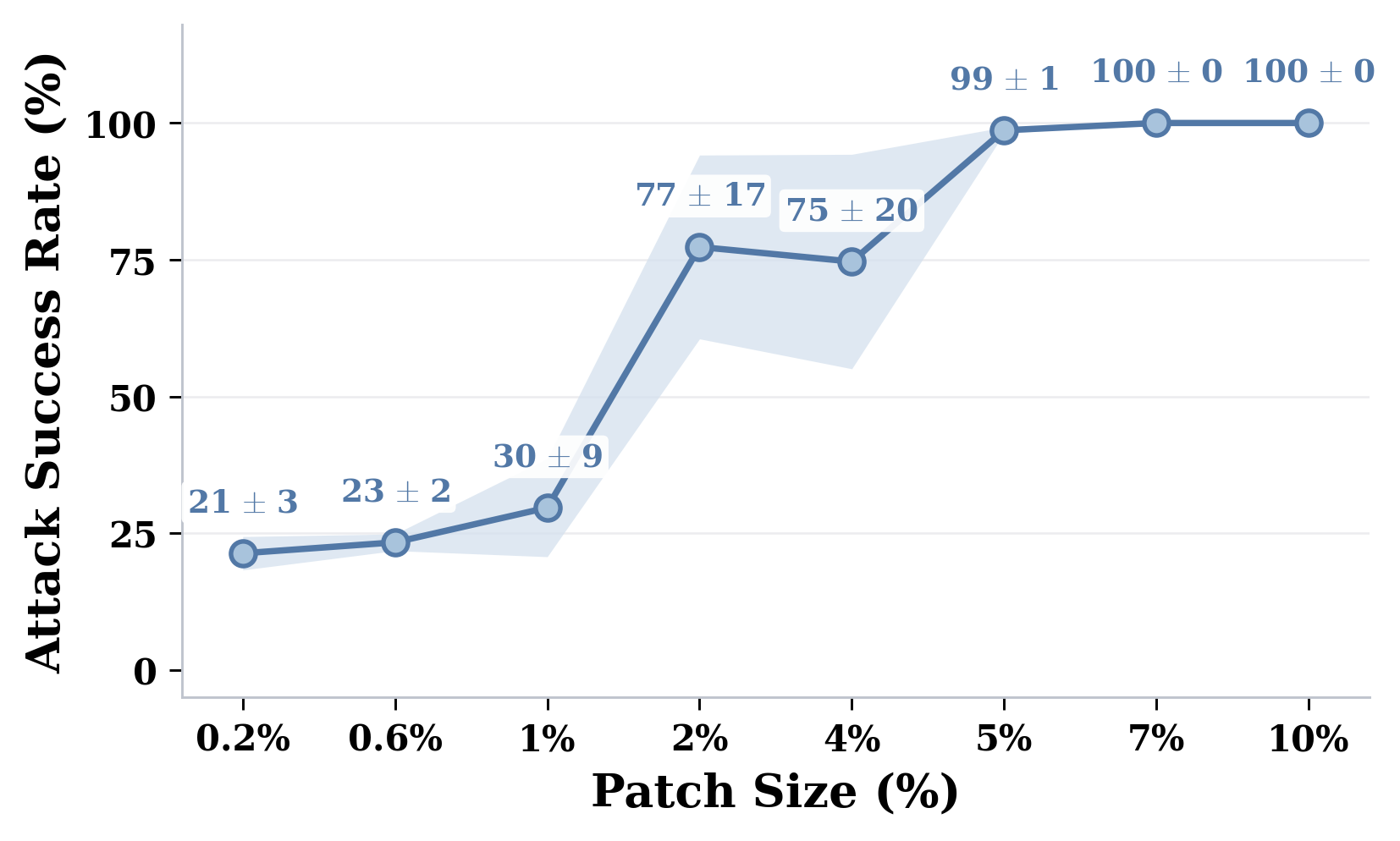}\\
% {\small Patch size}
\caption{\textbf{Patch size.}
White-box ASR versus patch size on LIBERO-Spatial; ASR rises past a size threshold
and saturates.}
\label{fig:Patch size}
\end{figure}

 Figure~\ref{fig:Patch size} studies the effect of patch size, measured by the fraction of image area occupied by the patch. We conduct this study on OpenVLA with the LIBERO under white-box access, and report the mean ASR over three random seeds with the other settings fixed as in Section:Experiments. Even with a small visual footprint, DURA already induces effective attacks. The ASR stays modest when the patch occupies at most $1\%$ of the image, but rises sharply to $77\%$ at only $2\%$ area. As the patch grows to $5\%$, the ASR reaches $99\%$ and then saturates at $100\%$ for larger ratios. We also observe that the variance across seeds is largest in the transition region (around $2\%$--$3\%$ area)
and vanishes once the patch saturates, where all three seeds reach $100\%$. These results show that DURA is effective with a low-area patch, while larger patches further improve attack stability.

\section{Quantitative Results AP}\label{app:AP}

  A higher AP indicates stronger steering toward the target action. For each suite we compare the no-patch \emph{Clean} baseline against our diffusion-guided patch under the \emph{white-box} (gradient access) and \emph{black-box} threat models. Under white-box access our patch raises AP to $82.0\%$ on average ($61$--$99\%$ across suites), and under black-box access to $67.3\%$ ($48$-$84\%$), peaking on Object and Spatial. In contrast, the clean policy AP only $25.0\%$ of the time ($19$--$32\%$),confirming that the freezing is caused by the optimized adversarial patch rather than by patch insertion alone. Each bar is a single run over $100$ trials per suite ($10$ tasks $\times$ $10$ rollouts).
\begin{figure}[h]
\centering
\includegraphics[width=\columnwidth]{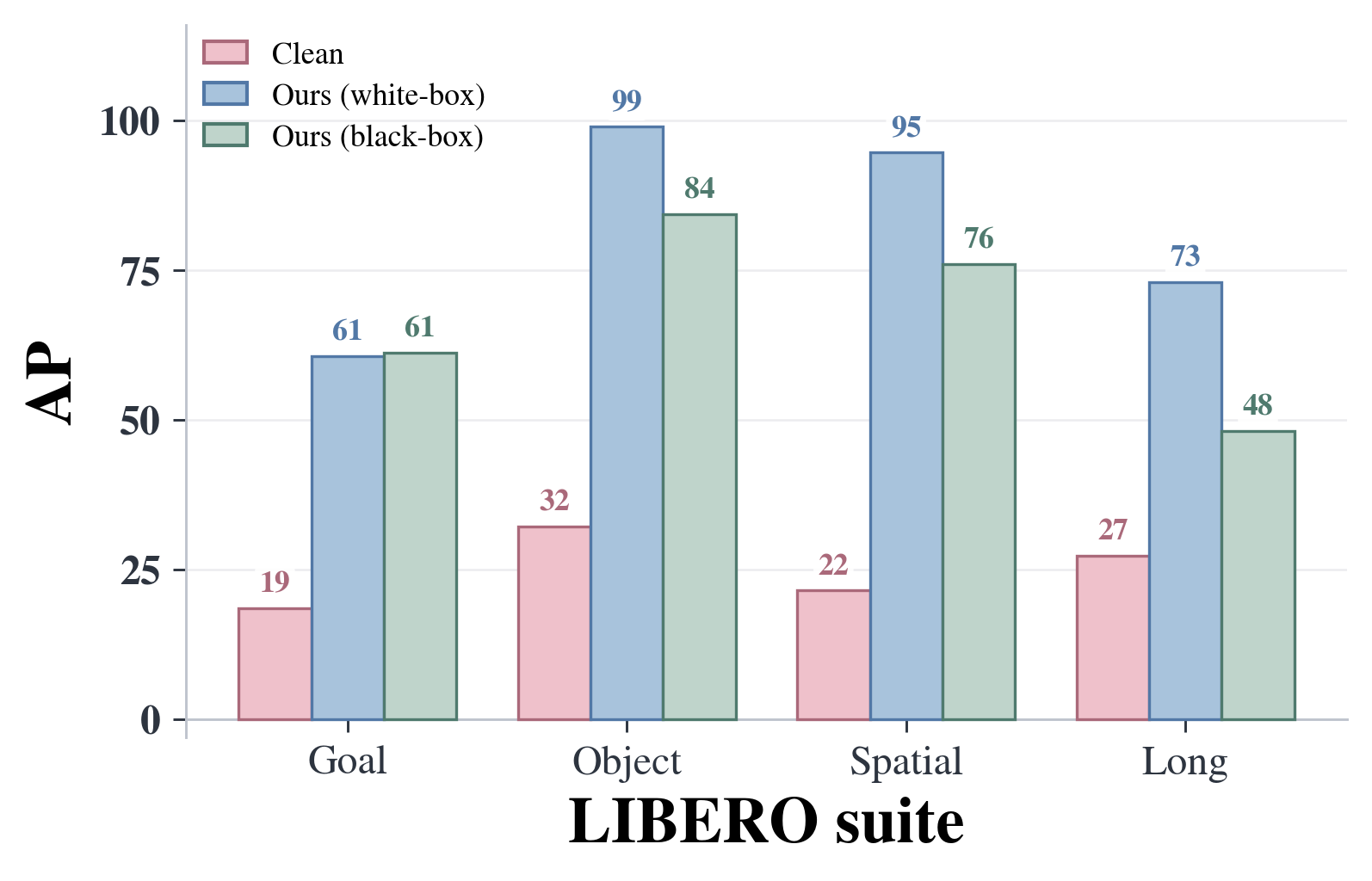}
  \caption{AP on OpenVLA across four LIBERO suites. Compared
  to the no-patch Clean baseline, our diffusion-guided patch sharply raises AP
  in both the white-box and black-box settings.}
  \label{fig:ap_openvla}
\end{figure}
\section{Visual Naturalness Evaluation Details}\label{app:natural}
We provide the evaluation details for the visual-quality analysis in Figure~\ref{fig:Visual naturalness}. For each attack method, we generate 8 patches and composite each patch into the same 10 LIBERO-Goal scenes, resulting in 80 patched crops per method. The no-patch reference uses the corresponding clean scene crops. All methods follow the same patch insertion location and image-compositing rule as in the main experiments.

To focus the evaluation on the patch and its nearby context, we crop a $192\times192$ local region from each scene (
$x\in[0,192),y\in[32,224)$
).
This crop contains the complete patch and the surrounding background. All metrics are computed on these local crops, and we report the mean and standard deviation over the evaluated crops.

We use both no-reference and reference-based metrics. Natural Image Quality Evaluator (NIQE)~\citep{mittal2012making} measures the deviation from natural image statistics without requiring a clean reference; lower values indicate more natural images. CLIP-Natural is computed using the CLIP-based image-quality assessment protocol~\citep{radford2021learning,wang2023CLIPIQA}, with the prompt pair "\emph{natural photo and synthetic photo}". Higher values indicate that the crop is more strongly aligned with natural photographic content. Before CLIP evaluation, each crop is resized to $224\times224$ and normalized to $[0,1]$.

For reference-based evaluation, each patched crop is paired with its corresponding no-patch crop. Deep Image Structure and Texture Similarity (DISTS)~\citep{ding2020image} measures perceptual structure and texture distortion, where lower values indicate smaller perceptual changes. Structural Similarity Index Measure (SSIM)~\citep{wang2004image} measures luminance, contrast, and structural consistency, where higher values indicate better preservation of the original scene.

We further include two local artifact measures. Boundary Seam Energy (BSE) measures the average CIE76 color difference~\citep{cie2019colorimetry} between adjacent pixels across the patch boundary; lower values indicate a smoother transition between the patch and the surrounding scene. Since the patch touches the left and bottom image borders, only the valid top and right boundaries are used. Patch Total Variation (Patch TV) measures the anisotropic total variation inside the patch region~\citep{rudin1992nonlinear}:
  \begin{equation}
 \mathrm{TV}(P)=
\mathbb{E}|P_{i+1,j}-P_{i,j}|
+
\mathbb{E}|P_{i,j+1}-P_{i,j}|.
  \end{equation}

Lower Patch TV indicates smoother patch texture and fewer high-frequency artifacts.
\section{Robustness to Input Transformations}\label{app:defense}
\begin{figure*}[t]
\centering
\includegraphics[width=\textwidth]{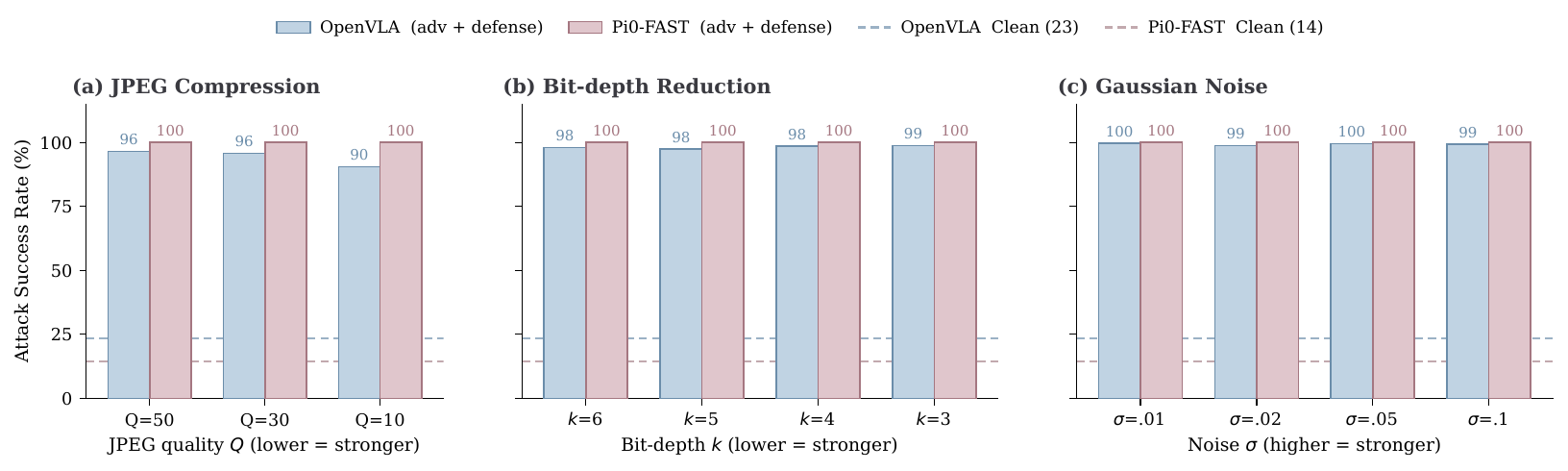}
\caption{\textbf{Robustness to input-transformation defenses (ASR).}
ASR (\%, $\uparrow$) averaged over the four LIBERO suites on OpenVLA and $\pi_0$-FAST under
JPEG compression, bit-depth reduction, and Gaussian noise at several strengths.}
\label{fig:defense_asr}
\end{figure*}
\begin{figure*}[t]
\centering
\includegraphics[width=\textwidth]{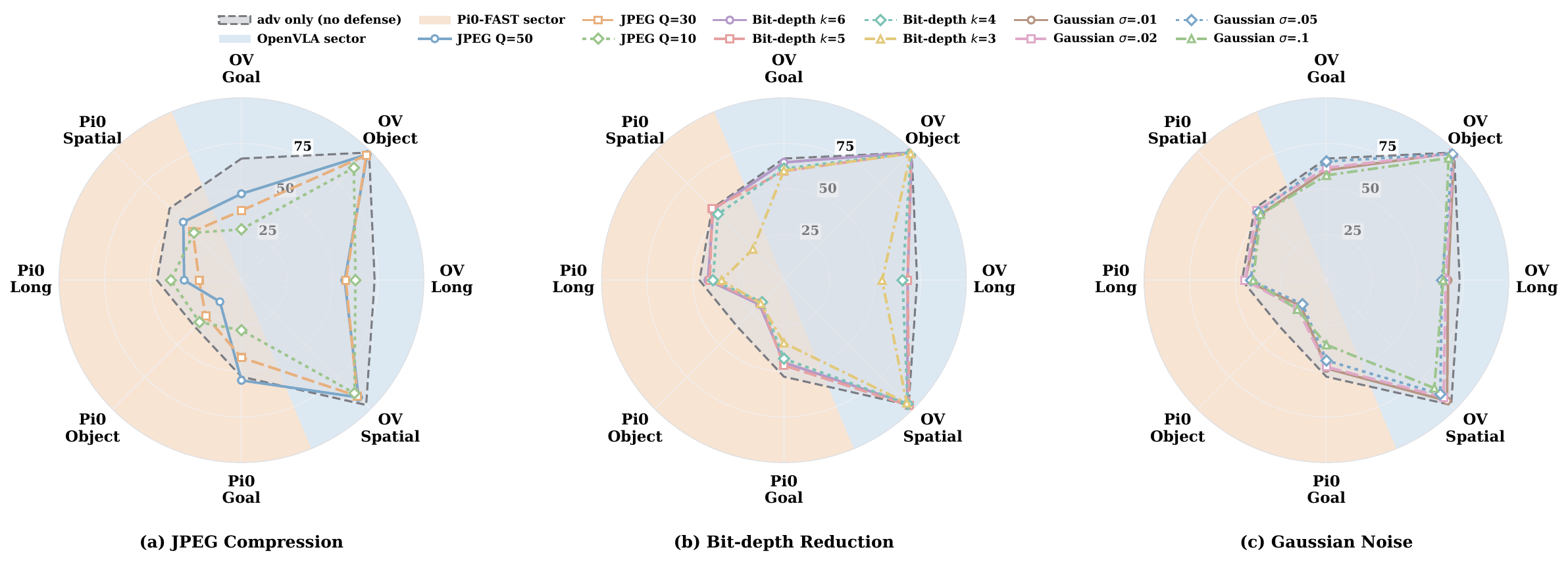}
\caption{\textbf{Robustness to input-transformation defenses (AP).}
AP per suite on OpenVLA(OV) and $\pi_0$-FAST under JPEG compression,
bit-depth reduction, and Gaussian noise at several strengths. The dashed line is the
no-defense attack, and each defense stays close to it.}
\label{fig:defense_ap}
\end{figure*}

% We evaluate common input-transformation defenses, including JPEG compression, bit-depth reduction, and Gaussian noise. As shown in Figure~\ref{fig:defense_ap}, DURA remains effective on both OpenVLA and $\pi_0$-FAST. Even under the strongest JPEG compression, the ASR remains 90\% on OpenVLA and 100\% on $\pi_0$-FAST, and it stays near 100\% under bit-depth reduction and Gaussian noise. These results show that DURA is robust to standard input transformations, suggesting that its adversarial effect is not easily removed by simple preprocessing.
We evaluate common input-transformation defenses, including JPEG compression, bit-depth
  reduction, and Gaussian noise, on both OpenVLA and $\pi_0$-FAST, and measure their effect
  with ASR (Figure~\ref{fig:defense_asr}) and per-suite AP (Figure~\ref{fig:defense_ap}).
  In terms of ASR, DURA stays effective under every defense and strength. Even under the
  strongest JPEG compression ($Q{=}10$), its ASR remains $90\%$ on OpenVLA and $100\%$ on
  $\pi_0$-FAST, and it stays at $98$--$100\%$ under bit-depth reduction and Gaussian noise,
  far above the clean baselines of $23\%$ and $14\%$.

  For AP, each radar axis in Figure~\ref{fig:defense_ap} is one LIBERO suite for OpenVLA
  (OV) and $\pi_0$-FAST (Pi0), and the dashed contour marks the no-defense attack. Across
  all three defenses, the AP contours stay close to this no-defense level, so the defenses
  rarely turn a targeted failure back into correct execution. The drop is largest under
  the strongest JPEG setting ($Q{=}10$) and bit-depth $k{=}3$, mostly on the OpenVLA
  suites, while Gaussian noise barely changes AP at any strength, and the $\pi_0$-FAST
  results stay close to no-defense throughout.

\end{document}